\documentclass[sigconf, nonacm]{acmart}

\usepackage{xcolor}         
\usepackage{listings}       
\usepackage{caption}
\usepackage[utf8]{inputenc}
\usepackage{subcaption}
\usepackage{cleveref}
\usepackage[ruled, vlined]{algorithm2e}

\definecolor{codegreen}{rgb}{0,0.6,0}
\definecolor{codegray}{rgb}{0.5,0.5,0.5}
\definecolor{codepurple}{rgb}{0.58,0,0.82}

\lstdefinelanguage{JavaScript}{
  keywords={async, await, break, case, catch, class, const, continue, debugger,
    default, delete, do, else, enum, export, extends, false, finally, for,
    function, if, import, in, instanceof, new, null, return, super, switch, this,
    throw, true, try, typeof, var, void, while, with, yield, with, as},
  keywordstyle=\color{blue}\bfseries,
  ndkeywordstyle=\color{darkgray}\bfseries,
  identifierstyle=\color{black},
  sensitive=false,
  comment=[l]{//},
  morecomment=[s]{/*}{*/},
  commentstyle=\color{purple}\ttfamily,
  stringstyle=\color{red}\ttfamily,
  morestring=[b]',
  morestring=[b]"
}

\lstdefinestyle{mypystyle}{
  language=Python,
  extendedchars=true,
  basicstyle=\footnotesize\ttfamily,
  showstringspaces=false,
  showspaces=false,
  numbers=left,
  numbersep=5pt,
  tabsize=2,
  breaklines=true,
  showtabs=false,
  captionpos=b,
  xleftmargin=0.5cm
}

\lstdefinestyle{mystyle}{
    commentstyle=\color{codegreen},
    keywordstyle=\color{magenta},
    numberstyle=\tiny\color{codegray},
    stringstyle=\color{codepurple},
    breakatwhitespace=false,
    breaklines=true,
    captionpos=b,
    keepspaces=true,
    numbers=left,
    numbersep=5pt,
    showspaces=false,
    showstringspaces=false,
    showtabs=false,
    tabsize=2
}

\AtBeginDocument{%
  \providecommand\BibTeX{{%
    \normalfont B\kern-0.5em{\scshape i\kern-0.25em b}\kern-0.8em\TeX}}}

\begin{document}

\title{A Static Analyzer for Detecting Tensor Shape Errors in Deep Neural Network Training Code}


\author{Ho Young Jhoo}
\affiliation{
  \institution{Seoul National University}
  \city{Seoul}
  \country{South Korea}}
\email{hoyoung.jhoo@sf.snu.ac.kr}

\author{Sehoon Kim}
\affiliation{
  \institution{Seoul National University}
  \city{Seoul}
  \country{South Korea}}
\email{shkim@ropas.snu.ac.kr}

\author{Woosung Song}
\affiliation{
  \institution{Seoul National University}
  \city{Seoul}
  \country{South Korea}}
\email{lego0901@gmail.com}

\author{Kyuyeon Park}
\affiliation{%
  \institution{Seoul National University}
  \city{Seoul}
  \country{South Korea}}
\email{kypark@ropas.snu.ac.kr}

\author{DongKwon Lee}
\affiliation{%
  \institution{Seoul National University}
  \city{Seoul}
  \country{South Korea}}
\email{dklee@ropas.snu.ac.kr}

\author{Kwangkeun Yi}
\affiliation{%
  \institution{Seoul National University}
  \city{Seoul}
  \country{South Korea}}
\email{kwang@ropas.snu.ac.kr}

\renewcommand{\shortauthors}{Jhoo, et al.}
\newcommand{\w}[1]{\ensuremath{\textit{#1}}}
\newcommand{\rar}{\rightarrow}
\newcommand{\Rar}{\Rightarrow}

\begin{abstract}
  We present an automatic static analyzer PyTea that detects tensor-shape errors in PyTorch code. The tensor-shape error is critical in the deep neural net code; much of the training cost and intermediate results are to be lost once a tensor shape mismatch occurs in the midst of the training phase. Given the input PyTorch source, PyTea statically traces every possible execution path, collects tensor shape constraints required by the tensor operation sequence of the path, and decides if the constraints are unsatisfiable (hence a shape error can occur). PyTea's scalability and precision hinges on the characteristics of real-world PyTorch applications: the number of execution paths after PyTea's conservative pruning rarely explodes and loops are simple enough to be circumscribed by our symbolic abstraction.  We tested PyTea against the projects in the official PyTorch repository and some tensor-error code questioned in the StackOverflow. PyTea successfully detects tensor shape errors in these codes, each within a few seconds.
\end{abstract}





\maketitle

\section{Introduction}
\subsection{Our Goal}

{\itshape Tensor shape mismatch} is a critical bug in deep neural network machine learning applications. Training a neural network is an expensive process that intends to terminate only when it finishes processing a huge amount of data through a sequence of tensor operations. In the middle of this time-consuming training process, if the shape of an input datum failed to fit with a tensor operation, the whole process abruptly stops wasting the entire training cost spent thus far, losing the trained, if any, intermediate result.

Our goal is to automatically predict at compile-time such run-time tensor-shape mismatch errors in PyTorch neural network training code.

\subsection{Structure of PyTorch Programs}

\begin{figure*}[t]
  \centering
  \includegraphics[width=0.8\textwidth]{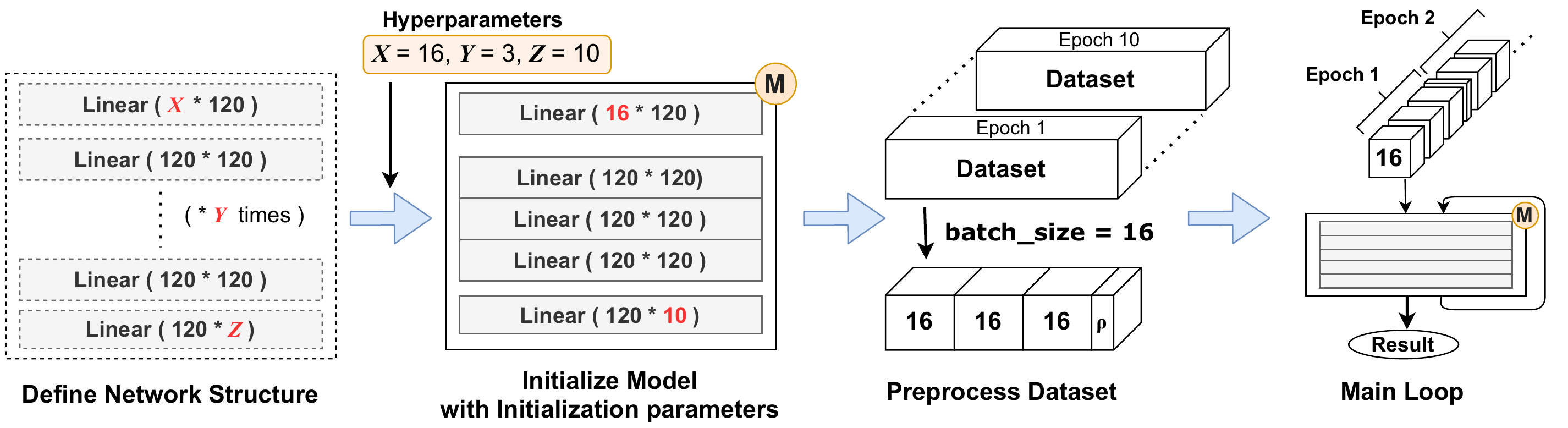}
  \caption{Typical structure of neural network training code in PyTorch.}
  \Description{It shows four parts of PyTorch code. Left to right: Define network structure, intialize model with initialization parameter, preprocess dataset, and run main loop.}
  \label{fig:teaser}
\end{figure*}

\label{sec:structure}

\begin{figure}[t]
  \centering
  \begin{lstlisting}[style=mypystyle]
## 1. DEFINE NETWORK STRUCTURE
class Net(nn.Module):
  def __init__(self, out_classes):
    super(Net, self).__init__()
    self.layers = nn.Sequential(
      nn.Linear(28 * 28, 120),
      nn.ReLU(),
      nn.Linear(120, out_classes)
    )

  def forward(self, x):
    x = x.reshape(x.shape[0], -1)
    x = self.layers(x)
    return x

## 2. INITIALIZE MODEL
model = Net(out_classes=10)

## 3. PREPROCESS DATASET
data = dataset.MNIST('./data', train=True,
    transform=[ToTensor()])
loader = DataLoader(data, batch_size=16)

## 4. RUN MAIN LOOP
for epoch in range(10):
  for batch, label in loader:
    # model(batch) == model.forward(batch)
    output = model(batch)
    loss = F.nll_loss(output, label)
    loss.backward()\end{lstlisting}
  \caption{Basic PyTorch training code.}
  \label{fig:lb-basic}
  \Description[Basic PyTorch neural network]{Simplfied Python/PyTorch neural network code that shows the general structure of PyTorch training.}
\end{figure}

Contemporary machine learning frameworks such as PyTorch~\cite{pytorch}, TensorFlow~\cite{tensorflow}, and Keras~\cite{keras} use Python APIs to build neural networks. Training a neural network with such frameworks is mostly patterned after a standard procedure which is illustrated in Figure~\ref{fig:teaser}. Typical PyTorch neural network training code can be divided into four stages. Figure~\ref{fig:lb-basic} shows a code example, a simplified image classification code taken from the official PyTorch MNIST classification example~\cite{pytorch_example}. We first define the series of neural network layers and make them into a single neural network module. To correctly assemble the layers, the returned tensor of the former layer must satisfy the input requirements of the next layer. We will see those requirements from the next section. The network is instantiated with some initialization parameters called hyperparameter, e.g., the number of hidden layers. Next, the input dataset is preprocessed and adjusted to the requirements of the network. Every dataset is cut into smaller same-sized chunks (called minibatches) from this stage. Finally, the main loop starts, and the minibatches are sequentially fed to the network. One epoch means a single loop that an entire dataset is passed to the network, and the number of epochs (datasets) usually differs depending on the purpose and structure of the neural network. Including the number of epochs, the numbers of iterations in the training code are determined to be constants in most cases, except the main training loop which depends on the size of a dataset.

\subsection{Tensor Shape Errors}
\label{sec:tensor-error}

\begin{figure}[h]
  \begin{subfigure}[t]{\linewidth}
    \begin{lstlisting}[style=mypystyle]
class Net(nn.Module):
  def __init__(self):
    super(Net, self).__init__()
    self.layers = nn.Sequential(
      ## 'B' represents batch size
      ## [B x 784] * [784 x 120] -> [B x 120]
      nn.Linear(28 * 28, 120),
      ## [B x 120] -> [B x 120]
      nn.ReLU(),
      ## [B x 120] * [80 x 10] -> ERROR!
      nn.Linear(80, 10))\end{lstlisting}
    \vspace*{-.6em}
    \caption{Error on the network structure.}
    \label{fig:lb-struct}
    \Description[PyTorch error sample 1]{Network initialization code that contains dimension mismatch in matrix multiplication.}
  \end{subfigure}
  \begin{subfigure}[t]{\linewidth}
    \begin{lstlisting}[style=mypystyle]
class Net(nn.Module):
  def __init__(self, batch_size):
    self.batch_size = batch_size
    # ...
  def forward(self, x):
    x = x.reshape(self.batch_size, -1)
    # ...

## some models may require exact batch size
model = Net(batch_size=64)

## POTENTIAL_ERROR 1:
##    argument 'drop_last=True' is essential
loader = DataLoader(data, batch_size=64)
# loader = DataLoader(data, batch_size=64,
#                           drop_last=True)

for epoch in range(10):
  for batch, label in loader:
    out = model(batch)
    ## ERROR ON THE LAST MINIBATCH
    ##     last batch size: 32 (!= 64)
\end{lstlisting}
    \vspace*{-.6em}
    \caption{Error on the last minibatch.}
    \label{fig:lb-batch}
    \Description[PyTorch error sample 2]{Data feed loop that crashes on the last batch because of the wrong batch size.}
  \end{subfigure}
  \begin{subfigure}[t]{\linewidth}
    \begin{lstlisting}[style=mypystyle]
## POTENTIAL ERROR 2: channel size can be 3
img = PIL.Image.open('./image.png').resize([28, 28])
# img = img.convert('L')

## ERROR WHEN THE IMAGE IS RGB.
tensor = to_tensor(img).reshape(28 * 28)
out = model(tensor)
\end{lstlisting}
    \vspace*{-.6em}
    \caption{Insufficient data preprocessing.}
    \label{fig:lb-data}
    \Description[PyTorch error sample 3]{Insufficient data preprocessing which does not consider both RGB and monochrome images.}
  \end{subfigure}
  \caption{Various type of tensor shape errors.}
  \label{fig:example}
  \Description[PyTorch error samples]{Sample codes of various types of tensor shape errors.}
  \vspace*{-1em}
\end{figure}

\begin{figure*}[ht]
  \centering
  \def\svgwidth{0.8\textwidth}
  \scalebox{1}{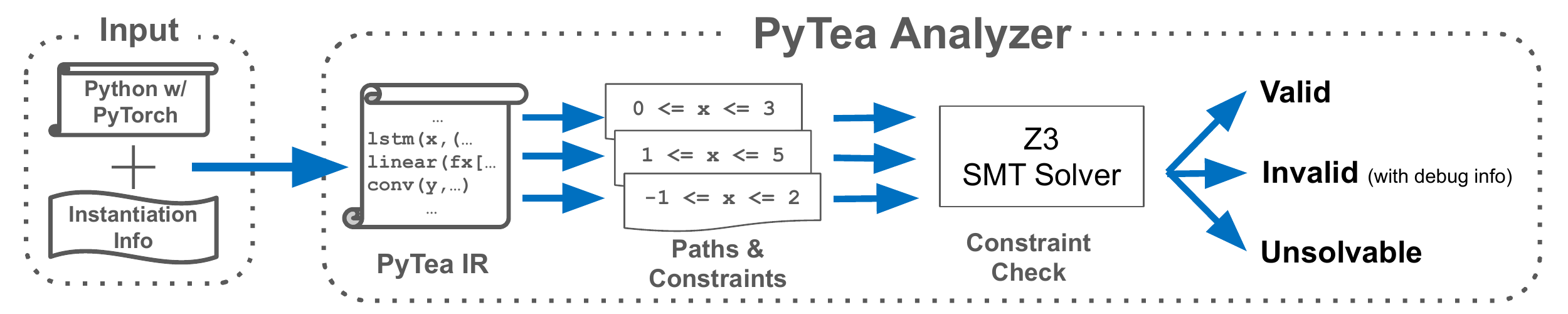}
  \caption{Overall architecture of PyTea.}
  \label{lb-arch}
  \Description[PyTea architecture]{Overall architecture of PyTea. It takes Python code and instantiation info (i.e., command-line arguments), translates into PyTea IR, extracts every paths and constraint sets, then feed them into SMT solver.}
\end{figure*}

\begin{figure}[ht]
  \centering
  \includegraphics[width=\linewidth]{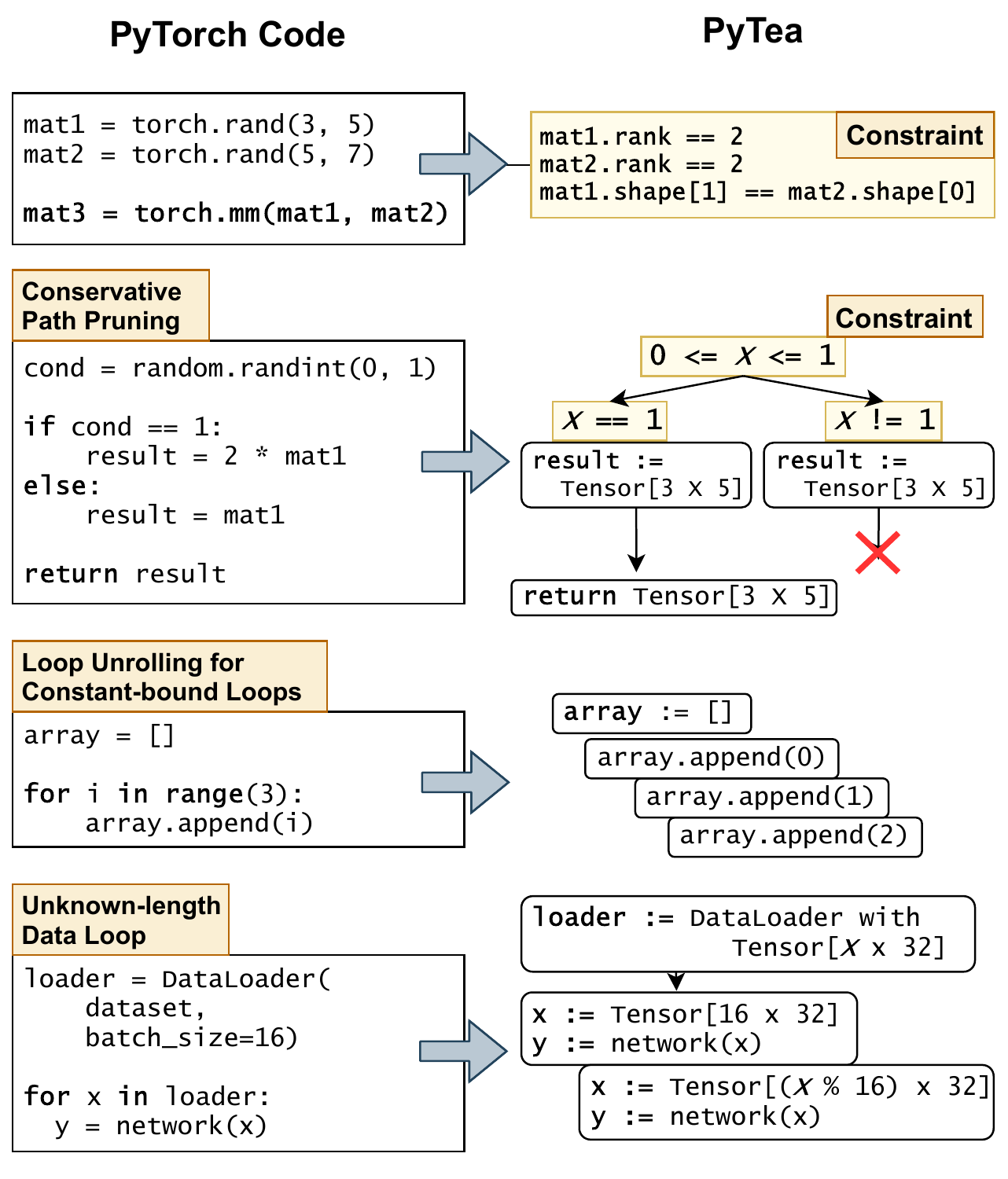}
  \caption{Constraint generation example.}
  \label{fig:const-gen}
  \Description[Constraint generation example]{Four strategies of path generations. Top to bottom: Cnstraint generation, Conservative path pruning, Loop unrolling for constant-bound loops, and Unknown-length data loop}
\end{figure}

\begin{figure}[ht]
  \centering
  \begin{lstlisting}[style=mypystyle]
class RandBlock(nn.Module):
  def __init__(self):
    super(RandBlock, self).__init__()
    self.layer = nn.Linear(32, 32)

  def forward(self, x):
    rand_num = random.randint(0, 1)

    if rand_num == 1:
      result = self.layer(x)
    else:
      result = x

    return result

model = nn.Sequential(
  [RandBlock() for _ in range(24)])
\end{lstlisting}
  \caption{Path explosion example.}
  \label{fig:explode}
  \Description[Path explosion example]{Neural network block which has runtime random variable in its feed-forward path.}
\end{figure}

Figure \ref{fig:example} presents the typical type of tensor shape errors, which are slight modifications of Figure~\ref{fig:lb-basic}. From the first example, the second {\verb|Linear|} layer (line 8), which multiplies the input with 80$\times$10-matrix, requires a specific shape of a tensor as an input. The first layer (line 6), however, returns a wrong-shaped tensor, and the overall pipeline will malfunction. This kind of error is called {\itshape tensor shape mismatch} error, simply, shape error.

Shape error is rather hard to manually find, only to be detected by running the program with an actual input. Indeed, the most notorious error for machine learning engineers is the error that can only be occurred after an immense amount of machine-hours.

\Cref{fig:lb-batch} shows another example. Its declaration of training data loader (line 14) hides a shape error. {\verb|DataLoader|} class slices the dataset sequentially by \texttt{batch\_size} and passes it to the model. If the total length of the dataset is not divisible by \texttt{batch\_size}, however, the size of the residual minibatch will be the non-zero remainder of the total length. See line 16: because the third parameter \texttt{drop\_last} is missing, the model assumes a consistent batch size (lines 10 and 6) hence the program will crash from the residual minibatch, losing the whole training hours. The recent massive networks like GPT-3~\cite{gpt3} require more than hundreds of machine-hours to train. This type of error must be noticed before its run.


\Cref{fig:lb-data} illustrates another shape error that can be arisen from a dataset, not a structure of the model. It does not take input from the pre-defined MNIST dataset but reads an image from a file. If the read image is RGB, which has 3$\times$H$\times$W dimensions, it will not fit into the reshape method that requires a tensor of 28$\times$28-elements. That means we have to convert it to a monochrome image before feeding it to the network. Even though it had been successfully tested with monochrome images, there can be a user who tests it with an RGB image, crashing the execution of the code.

Though several works \cite{ariadne, pythia, shapeflow, pytropos, semistatic} have reported tools to detect the shape mismatch errors of machine learning libraries, especially for TensorFlow \cite{tensorflow}, none of them have presented any static analysis tool that statically detects the shape errors for realistic Python ML applications. Real-world machine learning applications heavily utilize third-party libraries, external datasets, and configuration parameters, and handle their controls with subtle branch conditions and loops, but the existing tools still lack in supporting some of these elements and thus they fail to analyze even a simple ML application. To ensure that the shape error will not happen for {\itshape any} input data, we should statically infer a precise yet conservative range of each tensor shape and track its transformations through all possible execution paths.

\section{Overview of PyTea Analyzer}
\label{sec:overview}

To find out shape errors before runtime, we present a static analyzer {\bfseries PyTea} (PyTorch Tensor Error Analyzer). PyTea statically scans PyTorch applications and detects possible shape errors. PyTea analyzes full training and evaluation paths of the real-world Python/PyTorch applications with additional data processing and mixed usage of other libraries (e.g., Torchvision \cite{torchvision}, NumPy \cite{numpy}, PIL \cite{pillow})

Figure~\ref{lb-arch} illustrates the overall architecture of PyTea analyzer. It first translates the original Python codes into a kernel language, PyTea Internal Representation (PyTea IR). Then, it tracks every possible execution path of the translated IR and collects the constraints regarding tensor shapes that dictate the conditions for the code to run without a shape error. The collected constraint sets are given to Satisfiability Modulo Theories (SMT) solver Z3 \cite{z3} to judge that those constraints are satisfiable for every possible input shape. Following the result of the solver, PyTea concludes which path contains a shape error or not. If the constraint-solving by Z3 takes too much time, PyTea stops and tells "don't know".

\subsection{Assumptions}
\label{sec:assumption}

Given the typical structure of PyTorch neural network training code (Section~\ref{sec:structure}), we assume for the PyTea's input the followings about the PyTorch deep neural network training code:

\begin{itemize}
  \item[\textbf{A1}] Other than the training or evaluation dataset, every input value required to execute the code is injected by command-line arguments.
  \item[\textbf{A2}] There is no infinite loop and recursion. We assume that every loop bound except for the datasets will be fixed to a constant.
  \item[\textbf{A3}] The unknown loop bound for the datasets is only for the size of each dataset in an epoch, and every iteration is either with a fixed-sized minibatch of the dataset or with a smaller, residual minibatch.
  \item[\textbf{A4}] We assume that string-manipulation expressions have no effect on tensor shapes.
\end{itemize}

These assumptions are based on our observations that most PyTorch networks and codes can be statically determined to fixed structures once we give precise command-line arguments. Real-world PyTorch applications mostly construct their structures by command-line arguments or external configuration files like JSON files. Therefore, PyTea chooses to analyze programs only with exact command-line arguments.

For a few networks that are not resolved to a single fixed structure, we consider all possible structures. The number of the possible structures is to be controlled by our path-pruning technique, and sometimes, for an inevitable case, by timeout.

\subsection{Handling path explosions}

The number of possible paths is exponential to the number of branches in sequence. For some complex neural networks, such path explosion is possible. For example, Neural Architecture Search~\cite{nas} or Networks with Stochastic Depth~\cite{stochastic} have branches inside the network themselves. Figure~\ref{fig:explode} shows a representative path explosion case that utilizes a runtime random variable. We can notice that the feed-forward function ({\verb|forward(self, x)|}) has two execution paths in its body. The final structure of the network is made with 24 same blocks (line 17), which makes 16M paths.

We handle this exponential cost blow-up by means of conservative path-pruning and simple-minded timeouts. If we can find that the result of the binding scope of that feed-forward function is pure (i.e., do not change any global value), and its bounded value is indeed equal for every path and not related with the branch conditions, we then safely ignore other paths except for one. If a path explosion arises even if using this method, we then use a timeout. See Section~\ref{path_explosion} for more details.

\subsection{Handling Loops}

For the loops in typical PyTorch neural network programs, as we discussed in Section~\ref{sec:structure} and accordingly assumed in Section~\ref{sec:assumption}, we do not need the full power of static analysis~\cite{staticanalysis}. PyTea unrolls constant-bound loops (Assumption A2 in Section~\ref{sec:assumption}) and analyzes their straight-line code version.

For the unknown-bound loops for datasets, PyTea analyzes the loop body for just two cases with the aforementioned assumption A3. One is for the loop with a fixed-sized regular minibatch of an epoch. The other is for the loop with the residual minibatch. For example, see code in Figure~\ref{fig:const-gen}. For the third code box of Figure~\ref{fig:const-gen}, we can unroll the loop expression to 3 same expressions. If we do not know the length of the dataset, such as the fourth code box of Figure~\ref{fig:const-gen}, we use assumption A3 and consider only two cases for the two different sizes of minibatches.

\section{Analysis Steps}

\subsection{PyTea IR}

\begin{figure}[t]
  \[\def\arraystretch{1.4}
    \begin{array}{rrcl}
      \multicolumn{4}{l}{\w{Expression}}                                                                                            \\
       & \w{E}           & \rar & \w{n} \in \mathbb{Z} \, | \, \texttt{T} \, | \, \texttt{F} \, | \, \w{x} \,\,\, \text{(variable)} \\
       &                 & |    & \texttt{let} \, \w{x} \, \w{E} \, \w{E}                                                           \\
       &                 & |    & \texttt{if} \, \w{E} \, \w{E} \, \w{E}                                                            \\
       &                 & |    & \w{E} \,\, \w{bop} \,\, \w{E}                                                                     \\
       &                 & |    & \w{tensor-expr}                                                                                   \\
       & \w{bop}         & \rar & \w{numeric-op} \, | \, \w{compare-op}                                                             \\
       & \w{numeric-op}  & \rar & \texttt{+} \, | \, \texttt{-} \, | \, \texttt{*} \, | \, \cdots                                   \\
       & \w{compare-op}  & \rar & \texttt{<} \, | \, \texttt{=} \, | \, \cdots                                                      \\
       & \w{tensor-expr} & \rar & \texttt{mm} \, \w{E} \, \w{E}                                                                     \\
       &                 & |    & \texttt{reshape} \, \w{E} \, \w{E} \, \w{E}                                                       \\
       &                 & |    & \texttt{readImage} \,\, | \, \cdots                                                               \\
    \end{array}
  \]
  \caption{Abstract syntax of PyTea IR.\protect\footnotemark}
  \Description[Abstract syntax of PyTea IR]{Formal definitions of PyTea IR expressions}
  \label{fig:pytea-ir}
\end{figure}

\begin{figure}[ht]
  \centering
  \includegraphics[width=0.27\linewidth]{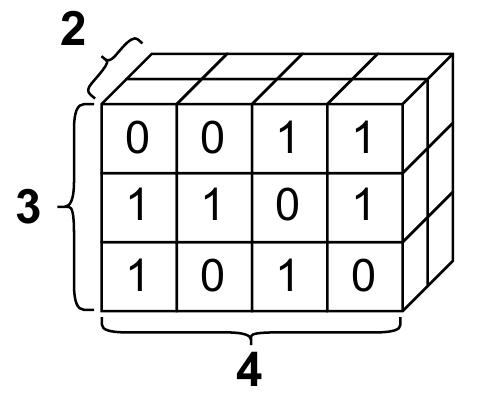}
  \caption{A tensor that has shape (2, 3, 4). The rank of this tensor is 3, and each dimension has size 2, 3, and 4.}
  \label{fig:cube}
  \Description[Tensor shape example]{Stacked cubes which aligned with X, Y, and Z-axis. Total 24 cubes.}
\end{figure}

\begin{figure}[t]
  \[\def\arraystretch{1.2}
    \begin{array}[t]{l}
      \begin{array}[t]{rclr}
        \multicolumn{4}{l}{\w{Constraint}}                                                                                                                \\
        c       & \rar & c \land c                                                        &                                                               \\
                & |    & c \lor c                                                         &                                                               \\
                & |    & \lnot \, c                                                       &                                                               \\
                & |    & e_b                                                              &                                                               \\
                & |    & e = e                                                            &                                                               \\
                & |    & e_n < e_n                                                        &                                                               \\
                & |    & \forall \alpha_n \in [e_n, e_n].c                                & \text{(} c \text{ is true forall integer } \alpha_n  \,\,\,\, \\
                &      &                                                                  & \text{in the interval)}                                       \\
        \multicolumn{4}{l}{\w{Value Expr}}                                                                                                                \\
        e       & \rar & e_s | \, e_n | \, e_b                                            & \text{(shape, number, or boolean)}                            \\
        \multicolumn{4}{l}{\w{Shape Expr}}                                                                                                                \\
        e_s     & \rar & \texttt{(} \, e_n \texttt{,} \cdots \, \texttt{,} e_n\texttt{)}  & \text{(tensor shape)}                                         \\
                & |    & \, \alpha_s                                                      & \text{(unknown shape)}                                        \\
                & |    & \, e_s \texttt{[} \, e_n\texttt{:}e_n\texttt{]}                  & \text{(shape slicing)}                                        \\
                & |    & \, e_s \texttt{@} \,e_s                                          & \text{(shape concat)}                                         \\
        \multicolumn{4}{l}{\w{Number Expr}}                                                                                                               \\
        e_n     & \rar & n                                                                & \text{(const number)}                                         \\
                & |    & \alpha_n                                                         & \text{(unknown number)}                                       \\
                & |    & e_n \, \w{bop} \,\, e_n                                          & \text{(binary operator)}                                      \\
                & |    & \texttt{rank} \,\texttt{(}\,e_s\texttt{)}                        & \text{(rank of shape)}                                        \\
                & |    & e_s\texttt{[}\,e_n\texttt{]}                                     & \text{(} e_n\text{-th dimension of shape } e_s \text{)}       \\
                & |    & \prod e_s                                                        & \text{(number of elements in} \,\,\,\,                        \\
                &      &                                                                  & \text{tensor of shape } e_s \text{)}                          \\
        \w{bop} & \rar & \texttt{+} \, | \, \texttt{-} \, | \, \texttt{*} \, |  \, \cdots &                                                               \\
        \multicolumn{4}{l}{\w{Boolean Expr}}                                                                                                              \\
        e_b     & \rar & \texttt{True} \,| \, \texttt{False}                              &                                                               \\
                & |    & \alpha_b                                                         & \text{(unknown boolean)}                                      \\
                & |    & e_{b} \land e_{b}                                                & \text{(conjunction)}                                          \\
                & |    & e_{b} \lor e_{b}                                                 & \text{(disjuction)}                                           \\
                & |    & \lnot \, e_b                                                     & \text{(negation)}                                             \\
                & |    & e = e                                                            & \text{(equality)}                                             \\
                & |    & e_{n} < e_{n}                                                    & \text{(less than)}                                            \\
      \end{array} \\
    \end{array}
  \]
  \caption{Abstract syntax of constraints.}
  \Description[Abstract syntax of constraints]{Formal definitions of PyTea IR constraints.}
  \label{fig:pytea-exp}
\end{figure}

\footnotetext{For the explanatory purpose, we did not include function calls and definitions. See supplementary material for detailed definitions of PyTea IR. Currently, we implemented 34 basic tensor expressions, and every other PyTorch API has been constructed with the basic expressions. The basic expressions are as following: \texttt{Torch.\_\_init\_\_}, \texttt{Torch.\_\_getitem\_\_}, \texttt{isSameShape}, \texttt{scalar}, \texttt{identity}, \texttt{broadcast}, \texttt{matmul}, \texttt{mm}, \texttt{bmm}, \texttt{item}, \texttt{repeat}, \texttt{expand}, \texttt{expand\_as}, \texttt{transpose}, \texttt{reduce}, \texttt{topk}, \texttt{view}, \texttt{conv2d}, \texttt{conv\_transpose2d}, \texttt{pool2d}, \texttt{batchnorm2d}, \texttt{cross\_entropy}, \texttt{cat}, \texttt{stack}, \texttt{unsqueeze}, \texttt{squeeze}, \texttt{diag}, \texttt{flatten}, \texttt{narrow}, \texttt{pixel\_shuffle}, \texttt{layer\_norm}, \texttt{pad}, \texttt{adaptive}, \texttt{interpolate}.}

As the first step of the analysis, the input Python code is translated into the kernel language, {\itshape PyTea IR}. See Figure~\ref{fig:pytea-ir}. PyTorch APIs are translated into tensor expressions that only define shape transformations, which PyTea IR focuses on.

The second step of the analysis is to scan the PyTea IR code and generate constraints.

\subsection{Constraint generation}

{\itshape Constraint}s are the conditions required by a PyTorch application so that it can be executed without any tensor shape error. For example, two operands of a matrix multiplication operation must share the same dimension. For each tensor operation ({\verb|mm|}, {\verb|reshape|}, {\verb|readImage|}, etc. of Figure~\ref{fig:pytea-ir}), the shape of the input tensor must obey the requirement of the corresponding operation.

Figure~\ref{fig:pytea-exp} shows the abstract syntax of the constraints. {\itshape Value expression} represents the value of PyTea IR expressions, which can be used inside shape constraints. When PyTea analyzes a PyTea IR, it traces tensor shapes and primitive values of Python and constructs symbolic value expressions. {\itshape Shape expression} represents the shape of tensors, which is basically a tuple of integers $(e_n, \dots, e_n)$. Figure~\ref{fig:cube} shows an example of a tensor with a shape $(2, 3, 4)$. Each integer is a dimension size. We call the number of dimensions as a {\itshape rank} of a shape. We can slice ($e_s \texttt{[}  e_n\texttt{:}e_n\texttt{]}$) a shape expression or concatenate ($e_s \texttt{@} \, e_s$) two shape expressions. For example, suppose a PyTea IR variable $\texttt{t}$ has shape $\texttt{(}2\texttt{,} 3\texttt{,} 4\texttt{)}$. Expression $\texttt{t[0]}$, which means the first sub-tensor of $\texttt{t}$ along the first axis, can be represented inside constraints as $\texttt{(}2\texttt{,} 3\texttt{,} 4\texttt{)[}1\texttt{:rank(t)]}$, or simply $\texttt{(}3\texttt{,} 4\texttt{)}$. In case of expression $\texttt{t}$'s shape is unknown($\alpha_s$), the shape of a sub-tensor $\texttt{t}\texttt{[}0\texttt{]}$ will be represented as $\alpha_s\texttt{[}1\texttt{:rank(}\alpha_s\texttt{)]}$.

\subsubsection{Constraint generation rules for PyTea IR}

To capture Python semantics and PyTorch shape transformations, PyTea follows the static semantics ($\sigma \vdash E : e, C $) of PyTea IR. Judgment ($\sigma \vdash E : e, C $) means that the PyTea IR expression $\w{E}$ is statically approximated by a symbolic value expression $e$ under environment $\sigma$ in case the constraint set $\w{C}$ ($\subseteq \w{Constraint}$) is satisfied. The environment $\sigma$ ($\in \w{Var} \overset{\w{fin}}{\rightarrow} \w{Value Expr}$) is a finite table that maps variables to symbolic value expressions.


The introduction of constraints happens for branch expressions or PyTorch APIs (See Section~\ref{sec:const-type}). The other expressions will collect constraints from their subexpressions. For example, for an add expression ($\w{E}_1 \texttt{+} \w{E}_2$), see: \[ \frac
  {
    \begin{array}{c}
      \sigma \vdash \w{E}_1 : e_n, \w{C}_1 \,\,\,\,\,\,\,\, \sigma \vdash \w{E}_2 : \w{e}_n', \w{C}_2 \\
    \end{array}
  }
  {
    \sigma \vdash \w{E}_1 \texttt{+} \w{E}_2 : \w{e}_n \w{+} \, \w{e}_n'  , \w{C}_1 \cup \w{C}_2
  }
\] The result value is symbolically ($e_n \w{+} \, e_n'$) where $e_n$ and $e_n'$ are symbolic results of $E_1$ and $E_2$ respectively. The result constraint set will be a union of the result constraint sets of $\w{E}_1$ and $\w{E}_2$.

Every symbolic variable originates from external input, e.g., random function or a dataset. Every expression in the constraints is constructed by these variables and constant values.

\subsubsection{Constraint types}
\label{sec:const-type}

In order to help the constraint resolution engine Z3 come up with a sensible counter-example that violates the derived constraints, we classify the constraints into two exclusive classes: {\itshape soft} and {\itshape hard} constraints. For Z3 to generate counter-examples, soft constraints can be violated, while hard constraints should not. Thus hard constraints are, for example, those from branch conditions or about the value range of the input. See Figure~\ref{fig:const-gen} again. Python built-in {\verb|random.randint|} function generates an unknown random variable within a given range {\verb|[0, 1]|}. We mark that bound constraint as a hard constraint. On the other hand, {\verb|torch.mm|} API demands that two input tensors have to be rank-2 ($x, y$) tensor and the second dimension ($y$-coordinate) of the first tensor have to be equal to the first dimension ($x$-coordinate) of the second tensor. This condition can be violated under the shape of the inputs, hence we mark it as a soft constraint.

\paragraph{Hard constraint generation}

Hard constraints are those for inputs and branch conditions. Input conditions restrict the initial ranges of each input. Branch conditions split each path into two.


Consider the following rule. \[
  \frac
  {
    \begin{array}{ll}
      c_1 = (1 \le \alpha_n \le 4) & (new \,\, \alpha_n)   \\
      c_2 = (0 < \alpha_n')        & (new \,\, \alpha_n')  \\
      c_3 = (0 < \alpha_n'')       & (new \,\, \alpha_n'') \\
      e_s = \texttt{(} \, \alpha_n \texttt{,} \alpha_n' \texttt{,} \alpha_n'' \texttt{)}
    \end{array}
  }
  {
    \sigma \vdash \texttt{readImage} : e_s, \{ c_1, c_2, c_3 \}
  }
\] The {\verb|readImage|} API is an image fetching API that creates a new 3-rank tensor which represents color channels, height, and width. The range of color channels is from 1 to 4, i.e., monochrome to RGBA, hence the constraint $c_1$ in the above rule. The symbolic value is a tensor of shape ($\alpha_n, \alpha_n', \alpha_n''$).

As another case, consider the following rule. \[
  \frac
  {
    \begin{array}{ll}
      \sigma \vdash E_1 : e_1, C_1   & \sigma \vdash E_2 : e_2, C_2 \\
      c = (e_1 \le \alpha_n \le e_2) & (new \,\, \alpha_n)
    \end{array}
  }
  {
    \sigma \vdash \texttt{randInt} \, \w{E}_1 \, \w{E}_2 : \alpha_n, \w{C}_1 \cup \w{C}_2 \cup \{ c \}
  }
\] The {\verb|randInt|} API generates a new random variable which is bound to given two numbers. This expression is used from the Python API {\verb|random.randint|}.

For branching case, see below: \[
  \frac
  {
    \begin{array}{c}
      \sigma \vdash E_1 : e_b, C_1 \,\,\,\,\,\,\,\, \sigma \vdash E_2 : e, C_2 \\
    \end{array}
  }
  {
    \sigma \vdash \texttt{if} \, \w{E}_1 \, \w{E}_2 \, \w{E}_3 : e, \w{C}_1 \cup \w{C}_2 \cup \{ e_b \}
  }
\]
\[
  \frac
  {
    \begin{array}{c}
      \sigma \vdash E_1 : e_b, C_1 \,\,\,\,\,\,\,\, \sigma \vdash E_3 : e, C_3 \\
    \end{array}
  }
  {
    \sigma \vdash \texttt{if} \, \w{E}_1 \, \w{E}_2 \, \w{E}_3 : e, \w{C}_1 \cup \w{C}_3 \cup \{ \lnot e_b \}
  }
\] The {\verb|if|} expression creates two paths depending on the branch condition $e_b$. If the branch condition can be evaluated to a constant boolean, we can safely drop one branch.

\paragraph{Soft constraint generation}

Soft constraints are the conditions with which PyTorch APIs must comply for them to run without a shape error. For instance, two operands of a matrix multiplication have to share the same middle dimension, and the reshape operation requires that the number of elements of the input tensor must be matched with the number of elements of the target shape. Each PyTorch API holds unique requirements of input conditions, and PyTea collects these requirements as soft constraints.

Following three rules, for example, PyTea collects such constraints from three representative APIs ({\verb|mm|}, {\verb|reshape|} and \\* {\verb|transpose|}): \[
  \frac
  {
    \begin{array}{c}
      \sigma \vdash \w{E}_1 : e_s, \w{C}_1 \,\,\,\,\,\,\,\, \sigma \vdash \w{E}_2 : e_s', \w{C}_2                                                                       \\
      \texttt{rank(} e_s \texttt{)} = \texttt{rank(} e_s' \texttt{)} = 2                                                                                                \\
      e_s'' = \texttt{(} e_s \texttt{[} 0 \texttt{]}, \, e_s' \texttt{[} 1 \texttt{])} \,\,\,\,\,\, c = ( e_s \texttt{[} 1 \texttt{]} =  e_s' \texttt{[} 0 \texttt{]} ) \\
    \end{array}
  }
  {
    \sigma \vdash \texttt{mm} \, \, \w{E}_1 \, \w{E}_2 : e_s'', \w{C}_1 \cup \w{C}_2 \cup \{ c \}
  }
\] The {\verb|mm|} API calculates a matrix multiplication of two 2-rank matrices. The second dimension of the first matrix must be equal to the first dimension of the second matrix following the basic rules of linear algebra.


The {\verb|reshape|} API redefines the shape of a tensor. Reshaping a tensor does not change or drop the value of a tensor, so the target shape must have the exactly same number of values as the original shape ($\prod e_s = \prod e_s'$): \[
  \frac
  {
    \begin{array}{l}
      \sigma \vdash \w{E}_1 : e_s, \w{C}_1  \,\,\,\,\,\,\,\,\,\,\,\,\,\,\,\,\,\,\,\,\,\,\, \sigma \vdash \w{E}_2 : e_n, \w{C}_2             \\
      \sigma \vdash \w{E}_3 : e_n', \w{C}_3 \,\,\,\,\,\,\,\,\,\,\,\,\,\,\,\,\,\,\,\,\,\,\, e_s' = \texttt{(} e_n \texttt{,} e_n' \texttt{)} \\
      c = ( 0 < e_n ) \land ( 0 < e_n' ) \land (\prod e_s = \prod e_s' )                                                                    \\
    \end{array}
  }
  {
    \sigma \vdash \texttt{reshape} \, \, \w{E}_1 \, \w{E}_2 \, \w{E}_3 : e_s', \w{C}_1 \cup \w{C}_2 \cup \w{C}_3 \cup \{ c \}
  }
\]

The {\verb|transpose|} API swaps two dimensions of the tensor $\w{E}_1$ along the $\w{E}_2$-axis and $\w{E}_3$-axis. Unlike the normal 2-rank matrix transposition, {\verb|transpose|} slices a tensor with $(\w{E}_2, \w{E}_3)$-plane and transposes each matrix on each cross-section: \[
  \frac
  {
    \begin{array}{c}
      \sigma \vdash \w{E}_1 : e_s, \w{C}_1  \,\,\,\,\,\, \sigma \vdash \w{E}_2 : e_n, \w{C}_2 \,\,\,\,\,\, \sigma \vdash \w{E}_3 : e_n', \w{C}_3 \\
      e_1 = e_s \texttt{[0:} e_n \texttt{] @ (} e_s \texttt{[} e_n' \texttt{])}                                                                           \,\,\,\,\,\,
      e_2 = e_s \texttt{[} e_n + 1 \texttt{:} e_n' \texttt{] @ (} e_s \texttt{[} e_n \texttt{])}                                                 \\
      e_s' = e_1 \texttt{ @ } e_2 \texttt{ @ } e_s \texttt{[} e_n' + 1 \texttt{:rank(} e_s \texttt{)]}                                           \\
      c = ( 0 \leq e_n < e_n' < \texttt{rank(} e_s \texttt{)})                                                                                   \\
    \end{array}
  }
  {
    \sigma \vdash \texttt{transpose} \, \, \w{E}_1 \, \w{E}_2 \, \w{E}_3 : e_s', \w{C}_1 \cup \w{C}_2 \cup \w{C}_3 \cup \{ c \}
  }
\] From this rule, we only consider the shape of the result, not the movement of the value inside the tensor.

\subsubsection{Handling path explosion}
\label{path_explosion}

Splitting execution paths whenever the analyzer encounters a branch can make the analysis cost grow exponentially. We can ignore some of them using the online constraint check, but we cannot for branches that use run-time input values.

However, we can still avoid path split if both paths behave identically in terms of tensor shape. The conservative conditions are as follows:

\begin{enumerate}
  \item Constraints collected from each path are not dependent on the branch condition, and
  \item Each path has no global side-effect, and
  \item Two paths' result symbolic values are the same.
\end{enumerate}

PyTea checks the above conditions locally, within the boundary of the $\texttt{let}$ expression containing each branch. When PyTea cannot statically decide on any of the three conditions, it safely assumes the conditions do not hold.

Most branches in PyTorch neural network blocks satisfy the above conditions. Typically, network blocks should result in a tensor with a fixed shape that matches with a requirement of the next block or the training target tensor. Those blocks' feed-forward path will be translated into nested $\texttt{let}$ blocks with branches that return the same-shaped tensor.

\subsection{Constraint check}

\subsubsection{Online constraint check}

To reduce the number of constraints and paths, our analyzer eagerly simplifies the symbolic expressions and constraints with primitive arithmetics and comparisons. By our eager, online constraint check, the ranges of each symbol can sometimes be known and be used to judge the subsequent constraints. If a branch condition can be simplified into constant true or false, we can trace only a single branch without splitting the path. If a constraint can be simplified to constant false, we can immediately report that the path is unsafe.

\begin{algorithm}[t]
  \caption{Offline Constraint Check with SMT Solver}
  \label{alg:solver}
  \Description{Algorithm for offline constraint check with SMT solver}
  \DontPrintSemicolon
  \SetKwProg{Fn}{Function}{:}{}
  \SetKwFunction{shortestStrings}{shortestStrings}
  \SetKwFunction{update}{update}
  \SetKwFunction{propagate}{propagate}
  \KwIn{$H, S$ - logical conjunctions of hard, and soft constraint sets}
  \KwOut{$\texttt{valid}$, $\texttt{invalid}$, $\texttt{dontknow}$, or $\texttt{unreachable}$}
  \Fn{$\texttt{analyze}(H, S)$} {
    \If{$\texttt{checkSat}(H) = \texttt{unsat}$} {
      \Return $\texttt{unreachable}$
    }
    \ElseIf{$S = \varnothing$}{
      \Return $\texttt{valid}$
    }
    $v = \texttt{checkSat}(\lnot (H \to S))$\;
    \If{$v = \texttt{unsat}$} {
      \Return $\texttt{valid}$
    }
    \ElseIf{$v = \texttt{sat}$} {
      \Return $\texttt{invalid}$
    }
    \lElse {
      \Return $\texttt{dontknow}$
    }
  }
\end{algorithm}

\subsubsection{Offline Constraint check}

\label{offline_step}

\begin{figure*}[ht]
  \centering
  \includegraphics[width=0.9\textwidth]{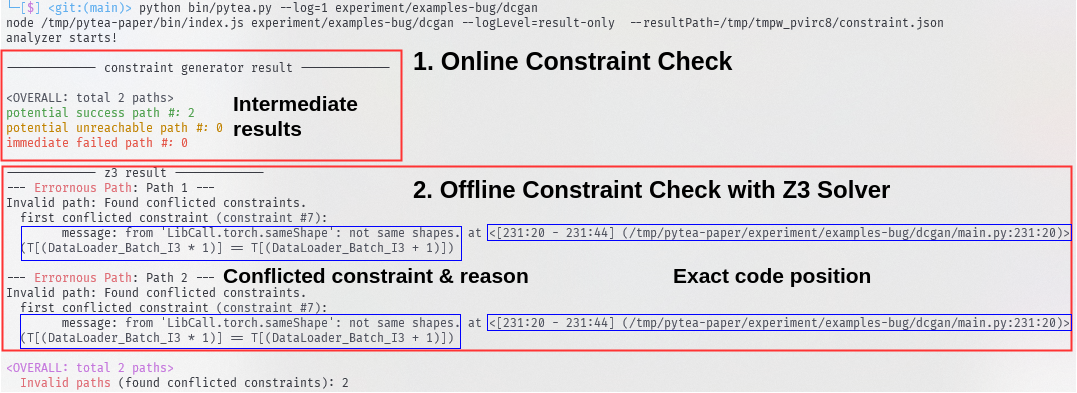}
  \caption{Test result of PyTea command-line tool.}
  \Description{Captured image of the result of PyTea command-line tool. PyTea analyzed errornous dcgan example and reported that the code has two invalid paths.}
  \label{fig:command}
\end{figure*}

PyTea feeds the collected constraints of each path to Z3. Algorithm~\ref{alg:solver} describes how we classify the Z3's result. The final result of PyTea analyzer can be divided into four cases:

\begin{itemize}
  \item Valid: Soft constraints are always satisfied under the hard constraints. It guarantees that shape error will not occur from this path.
  \item Invalid: A possible shape error is detected. There is a counterexample that makes soft constraints false under the hard constraints. We also report the generation position of the first broken constraint.
  \item Don't know: Z3 failed to decide whether constraints are satisfiable or not.
  \item Unreachable: There is a conflict between hard constraints in this path. In other words, it is impossible to reach this path under the given conditions. This can happen if a path had passed two contradicted branches.
\end{itemize}

If every path results in either unreachable or valid path, we can conclude that the input program has no tensor shape error.

\section{Evaluation}
\label{experiment}

Our experiments show PyTea's practical performance for real-world applications. To see the practicality of PyTea, we have collected several complete PyTorch applications and shape-related PyTorch bugs. First, we analyzed the official PyTorch example projects from GitHub repository pytorch/examples\cite{pytorch_example}. This repository consists 11 complete PyTorch applications about major machine learning tasks from Generative Adversarial Network (GAN)~\cite{dcgan} to Natural Language Processing. We also collected some PyTorch shape mismatch errors from StackOverflow and ran PyTea to statically detect them with PyTea. Finally, we conducted case analyses of several fully-functional, hand-made PyTorch applications such as Stochastic ResNet~\cite{stochastic}.

\paragraph{Experiment Settings}

PyTea analyzer is written in mainly TypeScript \cite{typescript}, and communicates with Python scripts to run Z3. We also used Pyright~\cite{pyright} to parse and track Python syntax. The experiments were conducted on R7 5800X CPU, node.js 16.0.0 with TypeScript 4.2.4, and Python 3.8.8 with Z3Py 4.8.10.0. We fixed the epoch size to 1 from the command-line arguments, but used default values for the other settings. We measured the total elapsed time from the cold boot to the termination of PyTea. The full options and codes are written in the supplementary materials. \footnotemark

\footnotetext{Link: \href{https://sf.snu.ac.kr/publications/pytea.zip}{{\itshape  https://sf.snu.ac.kr/pytea/}}}


\paragraph{PyTea command-line tool}

Figure~\ref{fig:command} shows an example snapshot of the analysis result of the PyTea command-line tool. It has analyzed one of the PyTorch example projects and prints the result of each phase of PyTea. It first prints out the online constraint check results and categorizes each path into three cases, potential success, potential unreachable, and immediate fail. The last one indicates that the online checker has found a constraint that can be false from that path. The potential unreachable path is the path which the online checker has found a false constraint, but there are certain unresolved branch conditions. That path will be checked at the next phase, and PyTea will examine whether the path has conflicted constraints only within the hard constraint set, which means that the path is unreachable from the beginning.

From the second step, PyTea delivers the collected constraint set of each path to Z3 solver and runs the offline constraint checks. The offline check will report the first conflicted constraint and its position of creation, i.e., the exact tensor expression or PyTorch API that causes an error. If the solver does not found any conflicted constraint, PyTea concludes that all the paths are valid, hence no tensor shape error is possible.

\subsection{Results}
\subsubsection{PyTea for PyTorch Examples}

\begin{table*}[ht]
  \caption{Analysis result of pytorch/examples code repository. The lines of library APIs encapsulated with the analyzer were counted separately. $\circ$: Analysis succeeded and found injected errors, $\triangle$: Analysis succeeded but requires a modification of the main code (e.g., provide explicit input tensor), $\times$: Failed to analyze.}
  \label{tbl-github}
  \begin{tabular}{lllll}
    \toprule
    Network                    & LOC (main + lib)  & PyTea       & Hattori et al. \cite{semistatic} & Total time (s) \\
    \midrule
    dcgan                      & 3714 (214 + 3500) & $\circ$     & $\times$                         & 1.75           \\
    fast\_neural\_style        & 4394 (338 + 4056) & $\circ$     & $\times$                         & 2.40           \\
    imagenet                   & 3820 (320 + 3500) & $\circ$     & $\times$                         & 2.40           \\
    mnist                      & 3607 (116 + 3491) & $\circ$     & $\times$                         & 1.59           \\
    mnist\_hogwild             & 3620 (129 + 3491) & $\circ$     & $\triangle$                      & 1.94           \\
    reinforcement\_learning    & 180 (180 + -)     & $\times$    & $\times$                         & -              \\
    super\_resolution          & 3886 (193 + 3693) & $\triangle$ & $\triangle$                      & 1.57           \\
    snli                       & 223 (223 + -)     & $\times$    & $\times$                         & -              \\
    time\_sequence\_prediction & 3333 (88  + 3245) & $\triangle$ & $\times$                         & 1.88           \\
    vae                        & 3593 (102 + 3491) & $\circ$     & $\triangle$                      & 1.70           \\
    word\_language\_model      & 3278 (361 + 2912) & $\triangle$ & $\times$                         & 1.81           \\
    \bottomrule
  \end{tabular}
\end{table*}

\begin{table}[ht]
  \caption{Analysis result of the StackOverflow questions. The numbers in parenthesis denote the URL id of each question.}
  \label{tbl-stack}
  \begin{tabular}{lllll}
    \toprule
    Question                                                       & PyTea   & Hattori et al. \cite{semistatic} \\
    \midrule
    Case 1 \href{https://stackoverflow.com/q/66995380}{(66995380)} & $\circ$ & $\times$                         \\
    Case 2 \href{https://stackoverflow.com/q/60121107}{(60121107)} & $\circ$ & $\times$                         \\
    Case 3 \href{https://stackoverflow.com/q/55124407}{(55124407)} & $\circ$ & $\times$                         \\
    Case 4 \href{https://stackoverflow.com/q/62157890}{(62157890)} & $\circ$ & $\times$                         \\
    Case 5 \href{https://stackoverflow.com/q/59108988}{(59108988)} & $\circ$ & $\times$                         \\
    Case 6 \href{https://stackoverflow.com/q/57534072}{(57534072)} & $\circ$ & $\times$                         \\
    \bottomrule
  \end{tabular}
\end{table}

For the experiment, we pass each project twice to the analyzer. For the first pass, PyTea analyzed the main code unmodified, and we check that PyTea does not inform false positives. Then, we injected artificial shape errors, which we subtract one from the first dimension of the target tensor, right before the neural network's loss calculation.

This simple method is decided on purpose. From this experiment, we focused on the speed of PyTea which shows the practicallity in order to be integrated to the code editor such as VSCode. This configuration can check the analysis time of the main network, and also confirm that PyTea tracks the tensor operations from the main network thoroughly, and we check PyTea does not report false negative results.

We have compared PyTea against another PyTorch analyzer of Hattori et al.~\cite{semistatic}. Table~\ref{tbl-github} shows the overall results. Among the 11 projects, PyTea successfully analyzed 6 projects without any modification of the original source code. For three projects with a complex data preprocessing stage, PyTea needs a bypass (i.e., code modification) of that stage to infer the shapes of input tensors. PyTea has also succeeded in finding these injected errors. As these results show, PyTea is quick and effective enough to be integrated into code editors. Meanwhile, Hattori et al.'s analyzer failed for almost all benchmarks. Furthermore, since their semi-static approach requires an explicit shape of the input tensor, we needed to feed them an exact network model and input tensors to compare its performance with PyTea.

Although we have aimed to analyze the codes without any modification, two projects are heavily dependent on third-party data managing libraries like {\verb|OpenAI-Gym|} \cite{gym}. Because, at the moment, we are focusing on the analysis of PyTorch-centered applications, we decided not to support those libraries for now. Supporting more libraries is straightforward and is our future work.

\subsubsection{PyTea for StackOverflow questions}

\begin{figure}[h]
  \centering
  \begin{lstlisting}[style=mypystyle]
class LSTM(nn.Module):
  def __init__(self, ...):
    # 7 lines...
  def forward(self, tokens):
    # 5 lines ...
    return out_scores

model = LSTM(embedding_matrix=np.zeros((1181, 100)))
loss_function = nn.NLLLoss()
optimizer = optim.Adam(model.parameters())

## CUSTOM INPUT
input = torch.ones(256, 4, dtype=torch.long)
target = torch.ones(256, 4, dtype=torch.long)
output = model(input)

## ORIGINAL
# output: [256 x 4 x 1181], target: [256 x 4]
#     SHAPE MISMATCH: [256 x 1181] != [256 x 4]
loss = loss_function(output, target)

## FIXED
# output: [1024 x 1181], target: [1024]
loss = loss_function(output.reshape(256*4, 1181), target.reshape(256*4))
\end{lstlisting}
  \caption{Example code of StackOverflow question. (Case 2)}
  \label{fig:lb-exam-stack}
\end{figure}

To show that PyTea can identify yet another set of real-world shape mismatches, we collected some PyTorch shape errors from StackOverflow questions. Recent TensorFlow analyzers~\cite{pythia,shapeflow} used a TensorFlow error dataset collected by Zhang et al. \cite{zhangbug}, but we manually gathered PyTorch shape mismatch cases rather than using their dataset, because of the fundamental difference of the structures between TensorFlow and PyTorch. We also considered porting the TensorFlow error dataset into PyTorch codes, but we concluded that the ported codes are fairly old and artificial and do not reflect the standard method to build a PyTorch application.

Table~\ref{tbl-stack} gives the analysis results of the 6 questions that we have collected. PyTea could detect every shape mismatch case from those questions. Following the analysis result, we could find the exact error positions and fix the shape mismatch cases. For example, the main code (Figure~\ref{fig:lb-exam-stack}) of Case 2 does not satisfy the shape conditions for the inputs of {\verb|NLLLoss|} (line 9). The {\verb|NLLLoss|} module requires that the shape of the first input tensor without the second dimension is equal to the shape of the second input tensor. PyTea found out that {\verb|NLLLoss|} could generate a shape error from our experiment. We then fixed the code according to the StackOverflow answer, and PyTea checked that every path became valid.

\subsection{Discovered Errors in PyTorch Applications}

We applied PyTea to several realistic PyTorch applications which contain potential shape errors or path explosion. PyTea-found shape errors include the typical type of shape errors that we introduced at Section~\ref{sec:tensor-error}. The complete projects and experiment scripts from this section will be in the supplementary material.

\subsubsection{Detecting insufficient data preprocessing}

We found a potential error at the data preprocessing stage from \texttt{fast\_neural\_style} application of pytorch/examples repository. As shown in Figure~\ref{lb-fast2}, \verb+Image.open+ does not guarantee the loaded image has channel 3, i.e., RGB image. Therefore, any training or inference stage with a monochrome image will fail if we miss the channel converting method like line 4. This error was remained from the initial version and was fixed by the latest commit \href{https://github.com/pytorch/examples/commit/a3f28a26851867b314f4471ec6ca1c2c048217f1}{(a3f28a2)} of the preprocessing script.

\subsubsection{Handling path explosion}

\begin{figure}[t]
  \centering
  \begin{lstlisting}[style=mypystyle]
def load_image(filename, size=None, scale=None):
  # POTENTIAL ERROR: channel size can be 1.
  img = Image.open(filename)
  # img = Image.open(filename).convert('RGB')
  # ...
  return img\end{lstlisting}
  \caption{Insufficient preprocssing of image file.}
  \label{lb-fast2}
\end{figure}

\begin{figure}[t]
  \centering
  \begin{lstlisting}[style=mypystyle]
def forward(self, x):
  residual = x

  if self.training:
    # sample random float value
    sample = self.m.sample().item()

    ### PATH EXPLOSION
    if sample > 0:
      out = self.conv1(x)
      out = self.bn1(out)
      out = self.relu1(out)
      out = self.conv2(out)
      out = self.bn2(out)

      if self.downsample is not None:
          residual = self.downsample(x)
      out = out + residual
    else:
      if self.downsample is not None:
          residual = self.downsample(x)
      out = residual
  # ...

  out = self.relu2(out)
  return out\end{lstlisting}
  \caption{Path explosion in Stochastic ResNet block.}
  \label{fig:sto-res}
\end{figure}

\begin{figure}[t]
  \centering
  \begin{tabular}{c}
    \begin{lstlisting}[style=mypystyle]
class NTXentLoss(torch.nn.Module):
  def __init__(self, batch_size, temperature):
    super(NTXentLoss, self).__init__()
    self.batch_size = batch_size
    # ...

  def forward(self, zis, zjs):
    batch = self.batch_size
    repr = torch.cat([zjs, zis], dim=0)
    sim = self.similarity_function(repr)

    ## zis: [B x N], sim: [2B x 2B]
    ## CONSTRAINT: -sim.shape[0] <= b <= sim.shape[0]
    l_pos = torch.diag(sim, b)

    # ...
    diag = torch.eye(2 * b)
    l1 = torch.diag(torch.ones(b), -b)
    l2 = torch.diag(torch.ones(b), b)
    mask = diag + l1 + l2
    mask = (1 - mask).type(torch.bool)
    # 'mask' tensor has (4b^2 - 4b) True values.

    negatives = sim[mask].view(2 * b, -1)
    # shape of 'negatives': (2b, 2b - 2)
    # ...

# ...
train_loader = DataLoader(
  train_dataset,
  batch_size=256,
  # drop_last=True, # ERROR
)
losses = train(net, train_loader)\end{lstlisting}
  \end{tabular}
  \caption{Shape inference which requires the exact values of a tensor.}
  \label{fig:lb-simclr}
\end{figure}

For a neural network model which contains a runtime path-explosion, PyTea analyzed it without a timeout. The \verb+stochastic-resnet+ example uses several deep learning techniques, mainly stochastic depth training~\cite{stochastic}. See Figure~\ref{fig:sto-res}. From this application, the building block of the network contains runtime branches (line 9) that can cause a path explosion. PyTea's path handling algorithm can successfully prune those branches and finishes without timeout. (Caveat: the overall data handling is somewhat hard to follow; we did not automatically reduce the repeat count of the main training loop. We explicitly reduced the length of the dataset (CIFAR-10) with a configuration file ({\verb|pyteaconfig.json|}), and without modifying the code itself.)

\subsubsection{Handling both regular and residual batch sizes in the training loop}

PyTea considers a residual minibatch in the training loop which leads to a shape error, as we discussed in Section~\ref{sec:overview}. We simplified the \verb+SimCLR+~\cite{simclr, simclr_repo} application to a single PyTorch-only script. From line 4 of Figure~\ref{fig:lb-simclr}, the main network class {\verb|NTXentLoss|} takes an exact batch size to initialize itself. So if we omit {\verb|drop_last|} parameter that removes the last batch at line 32, the last residual minibatch will lead to a crash if the total data size cannot be divided into the batch size. PyTea finds that the inequality between two batch sizes from line 14 of Figure~\ref{fig:lb-simclr} generates a shape error.

\subsection{Limitation of PyTea}

The main focus of PyTea is the detection of shape errors, so it does not perform general value analysis such as tracking the value of the tensor or array index out-of-bound exception.

If a shape of a tensor is dependent on the value of the other tensor, PyTea can miss a shape error. For instance, the \verb+view+ method at line 18 of Figure~\ref{fig:lb-simclr} requires that the element count of an input tensor is divisible by $2b$. Tensor masking by a boolean tensor (\verb+similarity_matrix[mask]+) returns a 1-D tensor whose length is equal to the number of \verb+True+ of the masking tensor. Although lines 10 to 14 guarantee that the masking tensor has $4b^2-4b$ \verb+True+, we do not know the \verb+view+ API will succeed since we do not track the exact value of a tensor.

\section{Related Works}

There is only one work~\cite{semistatic} of statically detecting shape mismatch of PyTorch applications. Hattori et al.~\cite{semistatic} presented a semi-static analysis of PyTorch applications that requires explicit tensor inputs. Because of the path-insensitive and semi-static approach, their tool is premature to fully statically analyze real-world applications. As shown in Table~\ref{tbl-github}, the performance of their tool is impractical.

For TensorFlow applications, the latest static analyzer is Pythia \cite{pythia}, following the same group's previous work Ariadne~\cite{ariadne}. Pythia is dependent on the Doop framework\cite{doop, ptaint} for Java pointer analysis and the Datalog language. Since Pythia's target is not Python, their coverage of Python and TensorFlow is still insufficient to handle real-world applications. For example, Pythia cannot analyze integer modular operation and tensor indexing and slicing, as shown in Figure~\ref{fig:pythia}. ShapeFlow~\cite{shapeflow} is a tester, a dynamic analyzer with fake TensorFlow libraries that only track shape transformations. Their dynamic approach achieved better performance and coverage than Ariadne and Pythia, but it requires a reduced dummy dataset to run their tool. It cannot detect a possibility of shape mismatch caused by an untested input dataset.

There are several works to solve the shape mismatch problems\cite{ariadne, pythia, shapeflow, pytropos, semistatic}, but they all have fundamental limitations to analyze PyTorch machine learning applications, such as the lack of support for handling external data, branches, and loops. Also, most of them work on TensorFlow applications.


Static analyses for Python programs have also been reported~\cite{fromherz, pytropos}. Notably, Cruz-Camacho's thesis\cite{pytropos} contains the shape analysis of NumPy\cite{numpy} array operators. However, their coverage of Python syntax is restricted that custom function and class declaration are not supported. PyExZ3\cite{pyexz3} is a value analyzer for Python language that implemented dynamic symbolic executor with Z3 backend. To port it for a shape mismatch problem needs a sizeable overhaul.

\section{Conclusion and Significance}

We have developed an automatic static analyzer PyTea that detects tensor-shape mismatch errors in PyTorch's deep neural network code. Our experiments have shown that PyTea's performance is practical in reality.

\paragraph{Significance} The tensor-shape mismatch error is a critical bug in deep neural net training code, yet hard to statically detect by programmers. We presented a solution to this problem: automatic static analyzer PyTea whose performance is realistic. The analysis design strikes a balance between the cost, accuracy, and coverage with a focus on the typical program structure of PyTorch deep neural net code base. PyTea is classified as a bug-finder, not a verifier: PyTea can have false positives or false negatives in principle, yet we observed no such cases in our experiments. PyTea is ready for public release.

\begin{figure}[ht]
  \centering
  \begin{lstlisting}[style=mypystyle]
import tensorflow as tf

one = 1
four = 1
if one == 1:
  four = 4

target = tf.ones((4, 5))

with tf.Session() as sess:
  t0 = tf.ones((3, 4))       # [3 x 4] * [4 x 5]
  p0 = tf.matmul(t0, target) # Pass: Correct

  t1 = tf.ones((3, 5))       # [3 x 5] * [4 x 5]
  p1 = tf.matmul(t1, target) # Error: Correct

  t2 = tf.ones((3, 5 % 2))   # [3 x 2] * [4 x 5]
  p2 = tf.matmul(t2, target) # Pass: False Negative

  t3 = tf.ones((3, 5))[0]    # [5] * [4 x 5]
  p3 = tf.matmul(t3, target) # Pass: False Negative

  t4 = tf.ones((3, 5))[0:1]  # [1 x 5] * [4 x 5]
  p4 = tf.matmul(t4, target) # Pass: False Negative

  t5 = tf.ones((3, four))    # [3 x 4] * [4 x 5]
  p5 = tf.matmul(t5, target) # Error: False Positive
  # ...\end{lstlisting}
  \caption{Basic tensor operations that Pythia~\cite{pythia} fail to analyze correctly.}
  \label{fig:pythia}
\end{figure}

\section*{ACKNOWLEDGMENT}

This work was partially supported by Korea Institute for Information \& Communications Technology Promotion (No.2021-0-00059), NAVER CLOVA (No. 0536-20200005) and Supreme Prosecutors' Office of the Republic of Korea (No. SPO2020A1103DIGITALB, No. SPO2021A1103DIGITALB). This work was also supported by BK21 FOUR Intelligence Computing(Dept. of Computer Science and Engineering, SNU) funded by National Research Foundation of Korea(NRF) (4199990214639).

\bibliographystyle{ACM-Reference-Format}
\bibliography{pytea}


\end{document}